\documentclass[lettersize,journal]{IEEEtran}
\usepackage{amsmath,amsfonts}
\usepackage{algorithmic}
\usepackage{algorithm}
\usepackage{array}
\usepackage[caption=false,font=normalsize,labelfont=sf,textfont=sf]{subfig}
\usepackage{textcomp}
\usepackage{stfloats}
\usepackage{orcidlink}
\usepackage{url}
\usepackage{verbatim}
\usepackage{graphicx}
\usepackage{cite}
\usepackage{booktabs}
\usepackage{multicol}
\usepackage{multirow}
\usepackage{kotex}
\usepackage{scalerel}
\usepackage{tikz}
\usepackage{hyperref}
\usetikzlibrary{svg.path}
\definecolor{orcidlogocol}{HTML}{A6CE39}
\tikzset{
  orcidlogo/.pic={
    \fill[orcidlogocol] svg{M256,128c0,70.7-57.3,128-128,128C57.3,256,0,198.7,0,128C0,57.3,57.3,0,128,0C198.7,0,256,57.3,256,128z};
    \fill[white] svg{M86.3,186.2H70.9V79.1h15.4v48.4V186.2z}
                 svg{M108.9,79.1h41.6c39.6,0,57,28.3,57,53.6c0,27.5-21.5,53.6-56.8,53.6h-41.8V79.1z M124.3,172.4h24.5c34.9,0,42.9-26.5,42.9-39.7c0-21.5-13.7-39.7-43.7-39.7h-23.7V172.4z}
                 svg{M88.7,56.8c0,5.5-4.5,10.1-10.1,10.1c-5.6,0-10.1-4.6-10.1-10.1c0-5.6,4.5-10.1,10.1-10.1C84.2,46.7,88.7,51.3,88.7,56.8z};
  }
}

\newcommand\orcidicon[1]{\href{https://orcid.org/#1}{\mbox{\scalerel*{
\begin{tikzpicture}[yscale=-1,transform shape]
\pic{orcidlogo};
\end{tikzpicture}
}{|}}}}

\usepackage{hyperref}

\hypersetup{
    colorlinks=false,
    linkbordercolor=white,
    urlbordercolor=white,
    pdfborder={0 0 0}
}

\begin{document}

\title{ECRC: Emotion-Causality Recognition in Korean Conversation for GCN}
\author{Jun Koo Lee$^{\textsuperscript{\orcidicon{0009-0001-1661-8249}}}$, \IEEEmembership{Student Member,~IEEE},
Moon-Hyun Kim$^{\textsuperscript{\orcidicon{0009-0001-2264-8620}}}$,
Tai-Myoung Chung$^{\textsuperscript{\orcidicon{0000-0002-7687-8114}}}$, \IEEEmembership{Senior Member,~IEEE}

\thanks{This research was supported by the Institute of Information \& communications Technology Planning \& Evaluation (IITP) grant funded by the Korean Government (MSIT) (No.2019-0-00421, AI Graduate School Support Program (Sungkyunkwan University)) and the Institute of Information \& Communications Technology Planning \& Evaluation (IITP) grant funded by the Korean Government (MSIT) (No.2020-0-00990, Platform Development and Proof of High Trust \& Low Latency Processing for Heterogeneous·Atypical·Large Scaled Data in 5G-IoT Environment)}

\thanks{Jun Koo Lee is with Department of Artificial Intelligence, College of Science, Sungkyunkwan University, Republic of Korea; (e-mail: \href{mailto:dlwnsrn0727@g.skku.edu}{dlwnsrn0727@g.skku.edu}).}

\thanks{Moon-Hyun Kim is with Department of Computing and Informatics, College of Science, Sungkyunkwan University, Republic of Korea; (e-mail: \href{mailto:mhkim@skku.edu}{mhkim@skku.edu}).}

\thanks{Tai-Myoung Chung is with Department of Computer Science and Engineering, College of Science, Sungkyunkwan University, Republic of Korea; (e-mail: \href{mailto:tmchung@skku.edu}{tmchung@skku.edu}).}}



\maketitle

\begin{abstract}
In this multi-task learning study on simultaneous analysis of emotions and their underlying causes in conversational contexts, deep neural network methods were employed to effectively process and train large labeled datasets. However, these approaches are typically limited to conducting context analyses across the entire corpus because they rely on one of the two methods: word- or sentence-level embedding. The former struggles with polysemy and homonyms, whereas the latter causes information loss when processing long sentences. In this study, we overcome the limitations of previous embeddings by utilizing both word- and sentence-level embeddings. Furthermore, we propose the emotion-causality recognition in conversation (ECRC) model, which is based on a novel graph structure, thereby leveraging the strengths of both embedding methods. This model uniquely integrates the bidirectional long short-term memory (Bi-LSTM) and graph neural network (GCN) models for Korean conversation analysis. Compared with models that rely solely on one embedding method, the proposed model effectively structures abstract concepts, such as language features and relationships, thereby minimizing information loss. To assess model performance, we compared the multi-task learning results of three deep neural network models with varying graph structures. Additionally, we evaluated the proposed model using Korean and English datasets. The experimental results show that the proposed model performs better in emotion and causality multi-task learning (74.62\% and 75.30\%, respectively) when node and edge characteristics are incorporated into the graph structure. Similar results were recorded for the Korean ECC and Wellness datasets (74.62\% and 73.44\%, respectively) with 71.35\% on the IEMOCAP English dataset.
\end{abstract}


\section{INTRODUCTION}

\begin{figure}[t!]
    \vskip 0.2in
    \begin{center}
        \centerline{\includegraphics[width=\columnwidth]{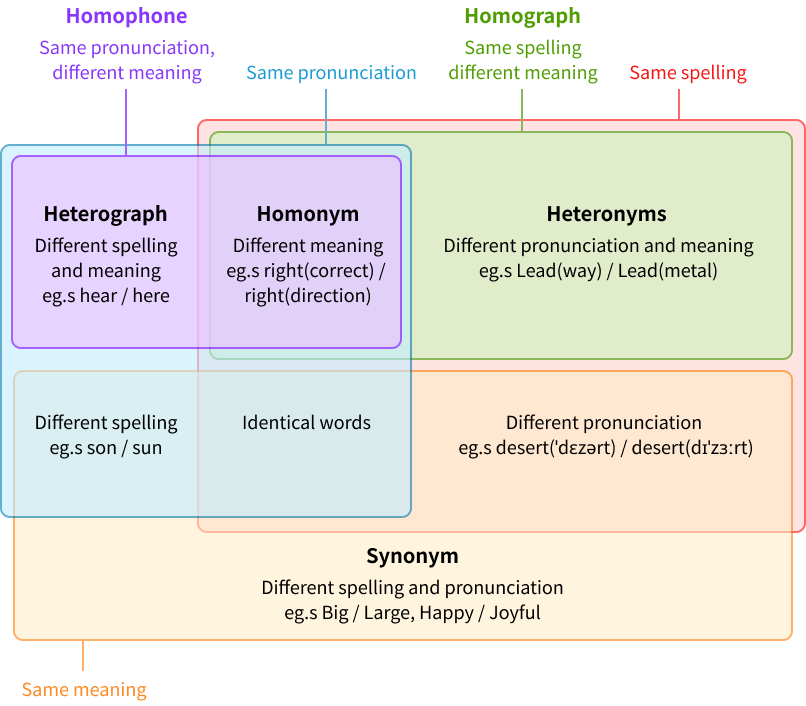}}
        \caption{Illustration of homonyms and homographs. Homophones refer to words that have the same pronunciation but different meanings, whereas homonyms are words that have multiple meanings but the same form.} 
    \end{center}
    \vskip -0.2in
\end{figure}

\IEEEPARstart{E}MOTION recognition in conversation (ERC) has achieved significant progress in the multi-task learning of emotion and causality in multi-turn dialogue datasets owing to the proliferation of social networking services (SNS) that allow the collection of diverse data types such as text, speech, and images. Embedding is a crucial preprocessing step in ERC and can be categorized into sentence-level embeddings [1],[2] and word-level embeddings [3],[4]. Word-level embedding of individual sentences is used in most embedding tasks. Previous research has focused on methods that consider the semantic combinations of words within sentences [5], learning embeddings related to specific emotions [6], and inherent characteristics of the words themselves [7].

Despite yielding meaningful results in various studies, the ERC word-embedding approach using single sentences has yet to address several issues. First, this approach struggles to capture contextual information in modeling the structural complexity of sentences because it considers the characteristics of individual words. Second, it has difficulty in handling cases where the same sentence carries multiple meanings and emotions, as illustrated in Fig. 1. In addition, this method encounters limitations in sentiment analysis owing to challenges in handling polysemy, semantic shifts, and relationships within sentences. Therefore, jointly incorporating sentence- and word-level embeddings more accurately reflects the interaction of words within sentences and the resulting meaning.

Several approaches in conversational data analysis have been explored to achieve high performance. These approaches consider the context of dialogue and personal characteristics of the speakers. DialogueRNN [17], which is one such approach employing attention-based recurrent neural networks (RNNs), is limited in terms of the attention mechanism and fails to consider the structure and relationships of emotions in conversations. Another approach is DialogueGCN [18], which represents dialogues as graph and model dependencies among speakers using graph convolutional neural networks (GCNs). However, it incurs a high computational cost during training and does not effectively derive the causes of emotions. DialogueCRN [19] has an inference module that extracts and integrates emotional cues through contextual reasoning networks (CRNs). However, there is insufficient evidence regarding the applicability of cognitive theories to emotion recognition, and these theories suffer from the complexity of the inference module, making it challenging to achieve generalizable performance. Therefore, to advance this field, research is required on methods capable of learning and modeling complex structures.

In this paper, we propose emotion-causality recognition in conversation (ECRC) with a novel graph structure. ECRC extends DialogueGCN [18] and EmoCaps [28] to represent complex sentence structures and multiple relationships. multi-task learning combines bidirectional long short-term memory (Bi-LSTM) with embeddings from  language models (ELMo) [1] and GCNs [20]. Multi-turn dialogue datasets are preprocessed using sentence- and word-level embeddings, introducing three different graph structures to enhance ECRC performance. For evaluation, we used Korean and English corpora labeled in the form of cause and causality.

\section{RELATED WORK}
\subsection{Emotion Recognition in Conversation}
Currently, emotion recognition in conversation (ERC) is being actively researched in natural language processing and text analysis. Recent studies have applied various models to enhance performance, especially by leveraging multi-modal data. These research endeavors and several related studies have primarily focused on constructing more complex model architectures based on deep learning to capture contextual complexity. Research has been actively conducted on various approaches, including  convolutional neural networks (CNN) and gated recurrent unit (GRU) networks to extract the main emotional features of context and contextual features, thereby applying an Attention mechanism to assign weights [24]. Additionally, studies have combined Bi-LSTM and Attention mechanisms to encode emotion-specific word embeddings in continuous representation of words [25]. There are also approaches that consider word order and context information through a combination of CNN and RNN [26]. These models offer the advantage of bidirectional learning, which enables focusing on important words, thereby delivering good performance on single sentences. However, they often have numerous hyperparameters, leading to longer training times, and their fixed structures make them less adept at handling complex sentence structures and multiple relationships; consequently, they perform poorly on multi-turn dialogue datasets.

Efforts have been made to address these issues. Khare et al. [27] systematically explored various embedding methods such as word2vec, global vectors (GloVe) for word representation, ELMo, and bidirectional encoder representations from transformers (BERT) using a multi-modal architecture. Seok [28] conducted a performance comparison between word- and sentence-level embeddings using Korean data. These studies highlight the need to consider various factors, including resource scalability and performance. Li et al. [29] developed the EmoCaps model, which extracts emotion vectors and forms emotion capsules by combining them with sentence vectors to yield classification results. Shen et al. [30] designed a directed acyclic graph network for conversational emotion recognition (DAG-ERC) by combining graphs and an RNN to structure the  history of conversations. These models define multiple components to integrate complex sentence structures and relationships into unified feature vectors.

\subsection{Emotion Cause Extraction}
In the field of emotion cause extraction (ECE), the initial extraction of specific emotions relies on language rules [8],[9] and machine learning algorithms [10]. However, this approach has two drawbacks [11]. First, emotions must be annotated before cause extraction, which limits the practical applicability to real scenarios. Second, the approach of annotating emotions and extracting causes ignores the fact that they are mutually referential. To address these issues, emotion–cause pair extraction (ECPE) was proposed, which integrates not only emotions but also causality. Research in the ECPE field includes two-stage approaches that perform individual emotion and cause extraction, followed by emotion-cause pair matching and filtering. Additionally, the research focusing on emotion prediction-oriented approaches (EPO-ECPE) [12] aimed to enhance the emotion–cause pair extraction by utilizing emotion prediction to the maximum extent. Furthermore, deep-learning-based models [13] for emotion–cause extraction have been explored, along with studies analyzing and modeling the interplay between emotions and causes from a linguistic perspective [14].

\section{Method}
The ECRC model utilizes ELMo for word embeddings instead of traditional word2vec models. While previous word2vec models calculated word embeddings in a fixed manner [15], [16], ELMo generates word embeddings by considering the diverse meanings and context within sentences [1]. This enables more accurate handling of polysemy and semantic changes based on context. Moreover, ELMo offers faster training times than Transformer-based models and has fewer hyperparameters to fine-tune. To represent the integrated feature vectors, we preprocessed the data into a graph structure with nodes and edges before training and evaluating the model using a GCN architecture. ECRC employs a three-step process to extract emotions and causality, as illustrated in Fig. 2. 

\begin{figure*}[t!]
    \vskip 0.2in
    \begin{center}
        \centerline{\includegraphics[width=\linewidth]{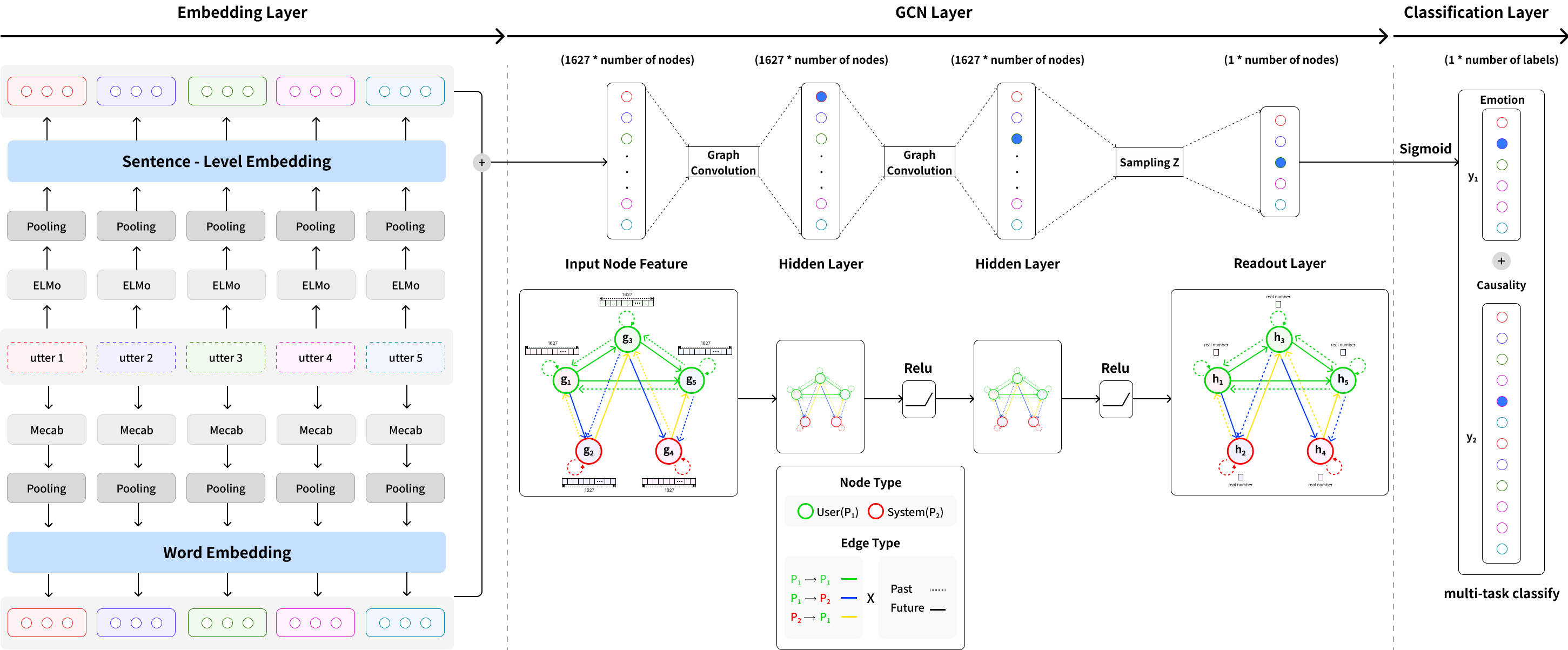}}
        \caption{Illustration of ECRC model structure. Initially, the embeddings are processed in the embedding layer. Then, in the GCN Layer, the processed data are transformed into a graph structure to simultaneously learn emotions and the causality that triggers them, for subsequent prediction of both classes. The GCN takes a graph structure with 1627 dimensions as input for training. In the hidden layer, information from neighboring nodes is combined to update the graph's hidden state. The readout step transforms the graph's features into a single vector, which serves as the output. This output predicts the label each class belongs to in the graph.}
    \end{center}
    \vskip -0.2in
\end{figure*}

\subsection{Embedding Layer(Step 1)}
In the embedding layer, sentence- and word-level embedding tasks are performed for a given corpus. A corpus consists of five utterances, and when embedding a sentence $u_{i} = (u_{1}, u_{2}, ..., u_{5})$, both Korean and English use the pretrained ELMo based on Bi-LSTM. The reason for structuring a corpus with five utterances is to ensure that the multi-turn dialogue ends with the user's final utterance, which has the most significant impact on the context. This arrangement is essential in facilitating a minimal multi-turn dialogue for context analysis. Adding more utterances to the corpus improves the performance.

\subsubsection*{\bf Bi-LSTM}
Assuming that the input utterance consists of N tokens ($t_{1}$, $t_{2}$, ...,$t_{n}$), each token $t_{k}$ represents a context-independent token for individual words. The forward LSTM models predict $t_{k}$ given $t_{1}$, $t_{2}$, ..., $t_{k-1}$ as input.

\begin{equation}
\label{eqn:label}
    p(t_{1},t_{2},...,t_{n}) = \prod_{k=1}^{N}p(t_{k}|t_{1},t_{2},...,t_{k-1}) \\
\end{equation}

Whereas the backward LSTM models predict $t_{k}$ when $t_{k+1}$, $t_{k+2}$, ...,$t_{N}$ are inputs.

\begin{equation}
\label{eqn:label}
    p(t_{1},t_{2},...,t_{n}) = \prod_{k=1}^{N}p(t_{k}|t_{k+1},t_{k+2},...,t_{N}) \\
\end{equation}

The bidirectional language model (biLM) combines the two LSTM models mentioned above to maximize the log-likelihood in both directions.

\begin{equation}
\label{eqn:label}
\begin{aligned}
    & \sum_{k=1}^{N} \left( \log p(t_{k}|t_{1},\ldots,t_{k-1};\Theta_{x},\overrightarrow{\Theta}_{LSTM},\Theta_{s}) \right. \\
    & \left. + \log p(t_{k}|t_{k+1},\ldots,t_{N};\Theta_{x},\overleftarrow{\Theta}_{LSTM},\Theta_{s}) \right)
\end{aligned}
\end{equation}

$\Theta_{x}$ represents the parameters for token representation $t_{1}$, $t_{2}$, ...,$t_{n}$  while $\Theta_{s}$ represents the parameters for the softmax layer. These two sets of parameters are shared in  both directions. However, the LSTM parameters have different values for the two LSTM models.

\subsubsection*{\bf ELMo}
ELMo employs a novel representation scheme, assuming the number of LSTM layers to be $L$. To achieve this, a total of $2L+1$ representations, including one input are concatenated

\begin{equation}
\label{eqn:label}
    R_{k} = \left\{x_{k},\overrightarrow{h}_{kj},\overleftarrow{h}_{kj}|j= 1,2,...,L, \right\} \\
\end{equation}

whre representation layer $j = 0$ and $L$ represent the forward and backward LSTM. Denoting $h_{kj}$ and $h_{kj}$'s concatenation as $h_{kj}$, the ELMo representation can be expressed using the following generalized formula:

\begin{equation}
\label{eqn:label}
    R_{k} = \left\{h_{kj}|j= 0,1,..,L, \right\} \\
\end{equation}

Finally, $R_{k}$ represents the concatenation of all representations of the $k$-th token, and is used to create the final ELMo embedding vector.

\begin{equation}
\label{eqn:label}
    ELMo_{k}^{task} = E\left ( R_{k};\Theta ^{task} \right ) = \gamma ^{task} \sum_{j=0}^{L}s_{j}^{task}h_{kj} \\
\end{equation}

All the outputs $h_{kj}$ from each LSTM layer are summed and multiplied by their respective softmax-normalized weights $s_{j}$. Finally, the result is multiplied by a scale parameter, denoted as $\gamma$.

\subsection{GCN Layer(Step 2)}
In previous semi-supervised research, there was a limitation in that connected nodes were likely to have the same label, and thus they could not capture additional information beyond similarity [20]. In the graph convolutional network (GCN) layer, embedding vectors are transferred into a graph structure to directly utilize the information of the connected nodes using the adjacency matrix from the utterances. We designed a weighted graph based on the information of connected nodes and conducted training and evaluation using the GCN model, thereby proposing the application of various node and edge features to enrich the graph structure. This framework is illustrated in Fig. 3.

\begin{figure}[t!]
    \vskip 0.2in
    \begin{center}
        \centerline{\includegraphics[width=\columnwidth]{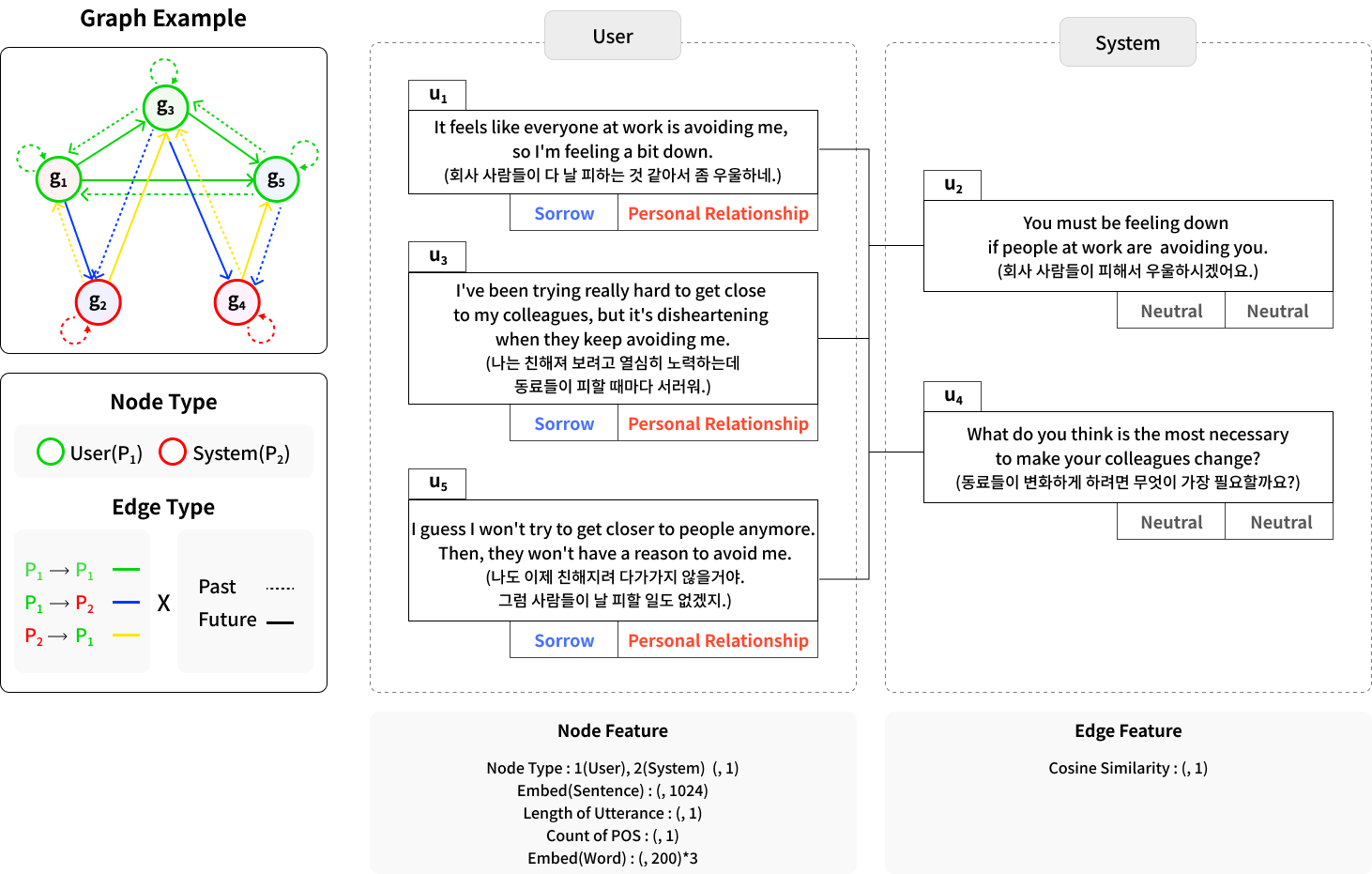}}
        \caption{Illustration of node and edge features. Each utterance is represented as $u_{i}$, and sequentially transformed into a graph structure $g_{i}$, where node and edge features are added. The dimensions of each node and edge feature are indicated, with the higher weight placed on the last sentence of the corpus, where labeling for emotion and causality is performed.}
    \end{center}
    \vskip -0.2in
\end{figure}

\subsubsection*{\bf Graph structure}
For the input utterance list $U = (u_{1}, u_{2},..., u_{N})$, where N is the number of input utterances, an undirected graph $G$ is constructed. $u_{i}$ represents utterance at $t$ = $i$; $U_{1}$ = $\{u_{2i+1} | i = 0, 1,...,\frac{N-1}{2}\}$ is a set of user utterances, while $U_{2}$ = $\{u_{2i} | i = 1, 2,...,\frac{N-1}{2}\}$ is a set of system utterances. $U = U_{1} \cup U_{2}$. The graph $G$ is represented as,

\begin{equation}
\label{eqn:label}
    G = (V,E,A,X) \\
\end{equation}

\begin{equation}
\label{eqn:label}
    X \in \mathbb{R}^{N \times M } \\
\end{equation}

Here $V$ is the set of nodes $V = \{V_{i} | i = 1, 2,..., N\}$, where $N$ is the number of nodes. Node $V_{i}$ corresponds to utterance $u_{i}$.

$E$ is the set of edges in the graph. The adjacent nodes $V_{i}$ and $V_{i+1}, i = 1, 2,..., N-1$ are connected. The graph has N-1 edges. Each user utterance node is connected to another user’s utterance node at different times. In other words, for $ \forall u_{j} \in U_{1}$, there is a connection to other $u_{k}$; $\forall u_{k} \in U_{1}$, $k \neq j$. whic has $\frac{N^{2}-1}{8}$ edges. Each system utterance node is not connected to other systems' utterance nodes. This is because it is unnecessary to predict emotions and causal relationships. The $\frac{N^{2} + 8N -9}{8}$ edges address the issue of increased computational costs in graphs with a large number of nodes. If each node is connected to every other node, then each node is connected by edges to all the other nodes, resulting in a total of $N^{2}$ edges.

$X$, $X \in \mathbb{R}^{N \times M}$ is a node feature matrix, where $M$ is the dimension of the node feature vector. $X_{i}$, the $i$-th row of $X$, denotes the feature vector of node $V_{i}$.

$X_{i}$ concatenates $\left[s_{i}, t_{i}, l_{i}, p_{i}, \omega_i^1, \omega_i^2, \omega_i^3\right]$. Here $s_{i} \in \mathbb{R}^{1024}$ is the sentence embedding vector generated by ELMo. Node type $t_{i} \in \left\{0,1\right\}$ categories the utterance $u_{i}$ as user utterance $\left(t_{i} = 1\right)$ or system utterance $\left(t_{i} = 0\right)$. 

$l_{i} \in \mathbb{R}$ is a sentence length which is computed as the number of tokens.

$p_{i} \in \mathbb{N}$ is a measure of diversity for part of speech($pos$), which is computed by counting the number of different $pos$. Especially in Korea, the diversity of $pos$ is closely related the to expression of emotions and causality; $pos$ is helpful because grammatical Korean modal expressions convey emotions and subjectivity[38].

$\omega_{i}^{j} \in \mathbb{R}^{200}, j = 1, 2, 3$ represents embedding vectors for three selected words in $u_{i}$. The top three words with the highest temr frequency-inverse documnet frequency(TF-IDF) values among the words in $u_{i}$ are selected

$A \in \mathbb{R}^{N \times N}$ is a symmetric adjacency matrix. For edge $E_{ij}$ connecting nodes $V_{i}$ and $V_{j}$, $A_{ij}$ is computed as the cosine similarity between two node feature vectors $X_{i}$ and $X_{j}$.

\begin{equation}
\label{eqn:label}
    A_{ij} = sim(X_{i}, X_{j}) = \frac{\sum_{k=1}^{M} X_{i}^{k} \times X_{j}^{k}}{\sqrt{\sum_{k=1}^{M}\left(X_i^k\right)^{2}} \sqrt{\sum_{k=1}^{M}\left(X_j^k\right)^{2}}}
\end{equation}

where $X_i^k$, $X_j^k$ are the $k$th elements of $X_{i}$ and $X_{j}$, respectively. An example of this is shown in Fig. 4.

\begin{figure}[htbp]
    \label{sample-table}
    \vskip 0.1in
    \centering
    \resizebox{0.5\textwidth}{!}{
        $A$ = 
        $\begin{bmatrix}
        0 & \mathrm{\textit{sim}}(x_2, x_1) & \mathrm{\textit{sim}}(x_3, x_1) & 0 & \mathrm{\textit{sim}}(x_5, x_1) \\
        \mathrm{\textit{sim}}(x_1, x_2) & 0 & \mathrm{\textit{sim}}(x_3, x_2) & 0 & 0 \\
        \mathrm{\textit{sim}}(x_1, x_3) & \mathrm{\textit{sim}}(x_2, x_3) & 0 & \mathrm{\textit{sim}}(x_4, x_3) & \mathrm{\textit{sim}}(x_5, x_3) \\
        0 & 0 & \mathrm{\textit{sim}}(x_3, x_4) & 0 & \mathrm{\textit{sim}}(x_5, x_4) \\
        \mathrm{\textit{sim}}(x_1, x_5) & 0 & \mathrm{\textit{sim}}(x_3, x_5) & \mathrm{\textit{sim}}(x_4, x_5) & 0
        \end{bmatrix}$%
    }
    \caption{Illustration of adjacency matrix of graph in Figure 3.}
    \vskip -0.1in
\end{figure}

\subsubsection*{\bf GCN}
In a typical GCN; $X$ and $A$ are passed to the next hidden layer using a hidden layer function $f$. If the $N \times M$-dimensional initial input $X$ is denoted as $H^{0}$, then the value of the $(l+1)$-th hidden layer can be defined using the hidden layer function f as follows.

\begin{equation}
\label{eqn:label}
    H^{(l+1)} = f(H^{l}, A) = \sigma(A H^{l} W ^{l})
\end{equation}

The hidden layer function $f$ can be expressed as a nonlinear function $\sigma $ that activates the hidden layer through the matrix multiplication of the adjacency matrix $A$, the $l$-th input matrix $H^{l}$, and the weight matrix $W^{l} \in \mathbb{R}^{M^{l} \times M^{(l+1)}}$, following the layer-wise relevance propagation rule. Similar to the number of filters in the CNN becoming the dimension of the feature map channel in the next layer, the matrix multiplication $H^{l} \times W^{l}$ also changes the dimension of the current layer's nodes from $M^{l}$ to $M^{(l+1)}$. Additionally, each row of the input matrix $H$ represents the node's feature, and by performing operations on all rows of nodes, weight sharing is achieved with each column of $W$. Subsequently, without considering locality, the adjacency matrix $A$ is multiplied to obtain a weighted sum that considers only the information of neighboring nodes.

However, this approach has three limitations. First, it does not consider self-loops. When the adjacency matrix $A$ is multiplied, the self-node is excluded. Therefore, to reflect the relationship between nodes, a self-loop must be added to the adjacency matrix. In other words, the identity matrix $I$ should be added to the adjacency matrix $A$ to allow the representation of a node to consider not only its relationship with other nodes, but also its own embedding. $\tilde{A} = A + I$ is the layer-wise relevance propagation rule with added self-loops expressed as follows;

\begin{equation}
\label{eqn:label}
\begin{aligned}
    & H^{(l+1)} =  \sigma (\tilde{A}H^{l}W^{l}) \\
\end{aligned}
\end{equation}

Second, using this method, nodes with many connections tend to have larger values. Due to the need to consider the relationships between nodes, the difference between nodes with a high degree of connected edges and those without can be significant. To mitigate this difference, data scaling is performed via normalization. In graph theory, adjacency matrix $A$ normalization involves multiplying the adjacency matrix by the inverse of the diagonal matrix D with degree values. The layer-wise relevance propagation rule with normalization is expressed as follows;

\begin{equation}
\label{eqn:label}
\begin{aligned}
    & H^{(l+1)} =  \sigma (D^{-\frac{1}{2}}\tilde{A}D^{-\frac{1}{2}}H^{l}W^{l}) \\
\end{aligned}
\end{equation}

\begin{itemize}
\item $H^{l}$ as the hidden state of $l$-th layer, where $H^{0}$ = X (the initial features of graph nodes).
\item $D$ where $D_{ii}$ = $\sum_{j}^{} \tilde{A}{ij}$, representing the diagonal matrix indicating the degree of each node.
\item $W^{l}$ as the weight matrix of the $l$-th layer.
\item $\sigma$ as the nonlinear function, where ReLU($\cdot$) was used.
\end{itemize}

Third, changing the positions of nodes can result in a change in the adjacency matrix $A$. To generalize $A$, a readout process is performed. The readout process transforms the feature matrix $H$ generated through the final convolution into a single vector to represent the entire graph, as follows.

The generated feature matrix $H^{L}$ through the final convolutional layer is converted to a feature vector $h^{l}$ by averaging the node feature vectors of $H^{L}$

\begin{equation}
\label{eqn:label}
\begin{aligned}
    & h^{l} = \frac{\sum_{i=1}^{N}H_{i}^{L}}{N}
\end{aligned}
\end{equation}

where $H_{i}^{L}$ is the $i$-th node feature vector of $H^{L}$

\subsection{Classification Layer(Step 3)}
In the classification layer, a classification model that predicts emotion and causality classes is created using a single vector transformed from the model output. In this process, the softmax function and the one-hot vector approach were utilized to calculate the probabilities for the emotion and causality classes. We use softmax because there is a dependency between the emotion and causality classes in the context of multi-task learning. The method is as follows.

Feature vector $h^{l}$ is applied to a fully connected layer to compare $z_{i}$ for each class i.

\begin{equation}
\label{eqn:label}
    z_{i} = \left ( W_{i}^{F} \right )^{T}h^{l}
\end{equation}

where $W_{i}^{F}$ is the weight vector of class $i$ of the fully connected layer.

\begin{equation}
\label{eqn:label}
    p_{i} = \frac{e^{z_{i}}}{\sum_{j=1}^{N_{c}} e^{z_{j}}}, i = 1,2,...,N_{c}
\end{equation}

$N_{c}$ represents the number of labels for both the emotion and causality classes, where emotion has six labels and causality has 12. Finally, the prediction for each class is determined by selecting the label with the highest probability. In this context, the actual values are represented using one-hot vectors. For example, when class $i$ is predicted, the corresponding one-hot vector, $y_{i}$, is represented as $y_{i}$ = [0, 0, ..., 1, ..., 0]. Here, one is placed at the position corresponding to label $i$, and all other elements are set to zero. This approach allows us to predict for emotion and causality classes using the softmax function and one-hot vector technique.

\section{Experiments}
In this section, the methodologies for the multi-task learning of emotions and causality in multi-turn dialogue datasets are extensively discussed using various models. The characteristics of sentences and their relationships are graphically represented in three different ways using the ECRC model for data training and evaluation. Additionally, to compare and analyze model performance, we employed the CNN, LSTM, and Bi-LSTM models, thereby demonstrating the superiority of the ECRC model in multi-task learning.

\subsection{Datasets}
As there is no specific Korean corpus available for ECRC, we labeled emotions and causality based on the AI-Hub ECC and Wellness data, whereby emotions were divided into six labels: 'Joy,’ 'Panic,’ 'Anger, ' ‘Frustration,’ 'Hurt,’ ‘Sorrow,' referring to emotional classification criteria in psychology [39],[40].

The causality category was divided into 12 labels based on various studies that analyzed the impact on emotions of environmental factors, such as social situations, cultural backgrounds, and family and friend relationships [43],[44]. These labels were: 'Career, Job,’ 'Personal relationship,’ ‘Love, Marriage, Childbirth,’ ‘Retirement,’ 'Financial,’ 'Disease, Death,’ 'Academic career,’ 'School violence/bullying', 'Working stress,’ ‘Personal relationships (couple, children),’ 'Family relationship,’ and ‘Health.'

In the context of causality, the impact of interpersonal relationships on emotions such as depression [45] was further classified, categorizing these into primary human relationships and secondary affiliations based on the factors contributing to the formation of interpersonal relationships. Primary human relationships are those formed by kinship, proximity, and academic affiliations, including parent-child, sibling, relative, and schoolmate relationships. Secondary relationships are established based on personal attractions, professional connections, and shared values (ideologies, beliefs, religions, and hobbies), including relationships with co-workers, lovers, friends, and members of social clubs [46]. 

We distinguish these relationships as personal. Additionally, based on research that categorized university students' anxiety factors into academic, job-related, interpersonal, and health-related domains [47],[48], these relationships were further divided into 'Academic career,' 'Career, job,' 'Personal Relationship,' and 'Health'; 'School violence/bullying' was included as a label based on previous studies [49],[50], which suggested that experiences of school violence/bullying could affect emotions. Labels like 'Financial, Retirement,’'Financial,' and 'Disease, Death' were derived from research highlighting the association between poverty, debt, and mental health issues, such as anxiety, depression, and suicide [51]. 'Working stress' was added based on previous studies [52], indicating emotional imbalances and changes in emotional expression due to job-related stress. Finally, 'Love, Marriage, Childbirth' was included as a label based on research examining the emotional impact of various social relationships [53],[54] and events deemed significant in social interactions, based on the comprehensive studies in [41] and [42] that discussed the elements constituting emotions and the factors that trigger them.

\subsubsection*{\bf Emotional Conversation Corpus(ECC)[21]}
The ECC was constructed using data from 1,500 individuals and comprised 15,700 speech utterances and 270,000 text corpus sentences. Each document was labeled for emotions and conversation scripts involving 2-3 people. In addition, labels for attributes such as age, gender, and physical condition were present but were removed during training, as they were not the primary parameters considered for emotion analysis. To align the datasets with others in terms of the number of classes, additional labels were added for emotions and causality. When a conversation script contains multiple emotions and situations, the label assigns more weight to the final utterance. The proportion of each label in the dataset is shown in Fig. 5.

\begin{figure}[t!]
    \vskip 0.2in
    \begin{center}
        \centerline{\includegraphics[width=\columnwidth]{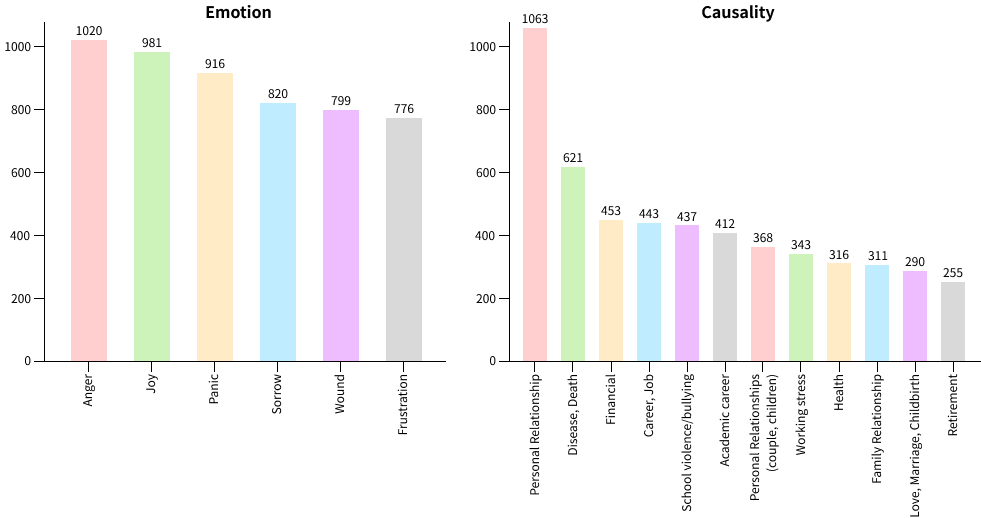}}
        \caption{Illustration of proportions of labels. This represents the labeling ratio of emotions and causality in the training data of the emotional conversation corpus, consisting of 5,312 samples. In the emotion category, 'Anger' accounts for 19.2\%, while in the causality category, 'Personal Relationship' has the highest proportion at 20.0\%. To address data bias, regularization and dropout techniques were applied during training.}
    \end{center}
    \vskip -0.2in
\end{figure}

\subsubsection*{\bf Wellness [22]}
Wellness consists of 5,232 user utterances and 1,023 chatbot responses, totaling 6,255 sentences created for 359 conversational intentions related to mental health counseling. The conversation scripts were segmented into sentence-level dialogues and categorized according to conversational intent, resulting in 12 major categories and 347 subcategories, including emotions and context. To ensure consistency with the other datasets, labels for emotions and causality were added.

\subsubsection*{\bf IEMOCAP [23]}
The IEMOCAP dataset contains daily conversations obtained from ten male and female participants, comprising 151 sessions and 6,803 utterances in audio and video formats. It includes the facial expressions and gestures of speakers, conversation scripts, and emotion labels in textual data. Only textual data were used for the training context analysis. The emotion labels were categorized into six classes: 'happy,' 'sad,' 'neutral,' 'angry,' 'excited,' and 'frustrated.'

\subsection{Preprocessing}

\subsubsection*{\bf Sentence Embedding}
All datasets were embedded with unique Korean tokens after removing English words, special words, and numbers. Padding was applied, whereby each utterance was padded to a maximum sentence length of 30 words to make input sentence lengths equal. The training/test dataset ratio was set to 8:2. The dataset sizes are listed in Table \uppercase\expandafter{\romannumeral1}. The input vectors and output results were transformed into one-hot vectors through one-hot encoding specific fine-tuning and the use of a bidirectional LSTM to capture diverse contexts, which combined make the method a strong alternative to BERT [15]. 

\begin{table}[htbp]
    \caption{The Training and Testing numbers for each dataset}
    \begin{center}
        \begin{tabular}{lccccr}
            \toprule
            Datasets & Train & Test \\
            \midrule
            ECC                  & 5313 & 1328\\
            Wellness             & 298 & 75\\
            IEMOCAP              & 679 & 170\\
            \bottomrule
        \end{tabular}
        \label{tab1}
    \end{center}
\end{table}

\subsubsection*{\bf Word Embedding}
 Word Embedding: Sentence-level embeddings aim to summarize all the information in a sentence into a single vector, which can result in the relative loss of structural characteristics of a sentence and alter relationships between words. To address this limitation, we applied word embedding to incorporate node and edge features. For word embedding, we used Mecab + word2vec for Korean and NLTK + word2vec for English. Because Korean and English have different language structures and grammatical features, morphological analyses must be performed accordingly. Korean is an agglutinative language that uses suffixes to convey form and grammatical information. By contrast, English is an inflectional language with relatively fixed word forms. Therefore, sentences with the same meaning often contain more speech parts in Korean. An example of this is shown in Fig. 6. 

\begin{figure}[t!]
    \vskip 0.2in
    \begin{center}
        \centerline{\includegraphics[width=\columnwidth]{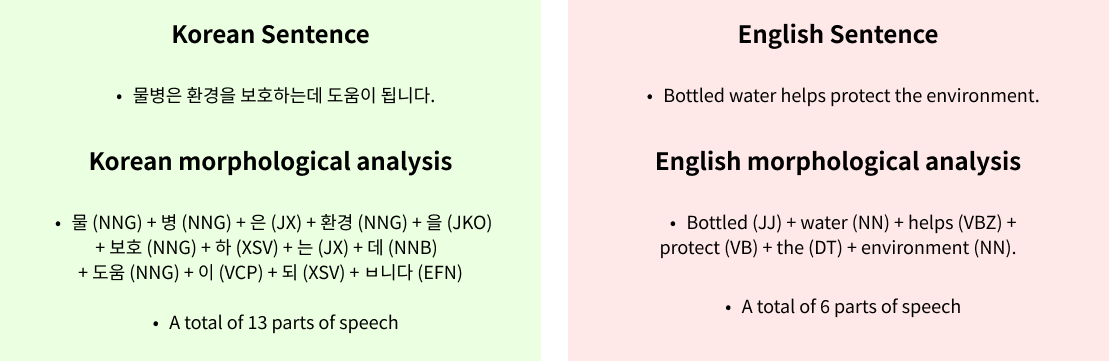}}
        \caption{Illustration of morphological analysis. Korean morphology is more complex than English due to structural and grammatical differences. In Korean, a single word can comprise multiple morphemes. For example, the sentence ‘물병은 환경을 보호하는데 도움이 됩니다' can be divided into 13 morphemes, while the equivalent English sentence 'Bottled water helps protect the environment' consists of only 6 morphemes. These characteristics were leveraged to gain a more precise understanding of context.}
    \end{center}
    \vskip -0.2in
\end{figure}

The use of various parts of speech allows for the creation of diverse sentence structures and more complex sentences. Therefore, we added sentence complexity as a node feature and analyzed based on the number of parts of speech. We further proposed a model to effectively incorporate this new structure.

\subsubsection*{\bf Mecab}
For the Korean morphological analysis, we used Eunjeon’s Mecab library [35]. Although other Korean morphological analyzers, such as Khaiii, Hannanum, Komoran, and Okt, are available, we chose Mecab because of its short analysis time. The results of this analysis are presented in Fig. 7 [37].

\begin{figure}[t!]
    \vskip 0.2in
    \begin{center}
        \centerline{\includegraphics[width=\columnwidth]{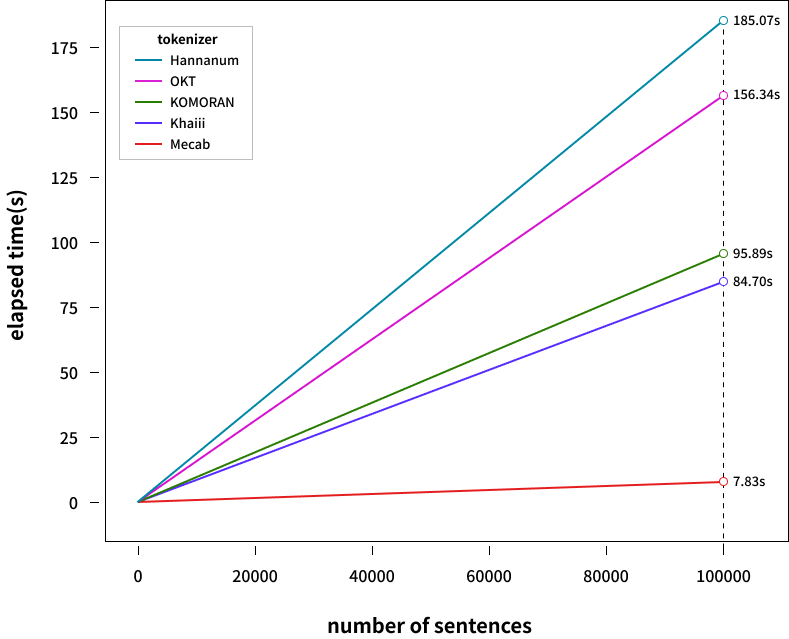}}
        \caption{Illustration of conversation data. The comparison of five Korean morphological analyzers reveals that Mecab  has the shortest analysis time. Mecab utilizes a morpheme dictionary to tokenize input text into words and morphemes in determining the part of speech for each morpheme. Mecab demonstrated advantages regarding speed and efficiency compared to other morphological analyzers, performing morphological analysis on 100,000 sentences in just 7.83 s. This performance is approximately 10.8 times faster than that of Khaiii, which is the second fastest at 84.70 s.}

    \end{center}
    \vskip -0.2in
\end{figure}

\subsubsection*{\bf word2vec}
Word embedding was performed using prediction-based word2vec [36], which represents similarity by vectorizesing words to represent similarity in a low-dimensional vector space.

\subsection{Model}
The model consisted of a three-layer MLP architecture with two hidden layers, each consisting of 128 hidden units and one output layer. The input dimension was 1627, representing a 5-node graph with 1627-dimensional vectors, and the output dimension was (6+12) for the predictions. Each MLP layer incorporated ReLU nonlinearity and dropout. The model was trained for 200 epochs with a batch size of 32. ECRC utilizes an undirected weighted graph, whereby graph G is created using the NetworkX library and augmented with node and edge attributes using the deep graph library (DGL).  The GCN model was selected because of its suitability for capturing complex structural contexts in conversational data, as opposed to linear contexts [17],[18]. We also drew inspiration from models such as FSS-GCN [31], EGAT [32], and ME-GCN [33], which enhance GCN's handling of node and edge features to improve classification performance. The initial learning rate was set to 0.001 and the model employed CrossEntropyLoss as the loss function. Model performance was evaluated by measuring precision, recall, and F1-score on a test dataset labeled for emotion and causality predictions. The following three graph-based models were utilized:

\begin{itemize}
\item \textbf{Case1 Sentence-Level Embedding Only}
- sentence-level embeddings were applied as node features using ELMo.
\item \textbf{Case2 Sentence-Level Embeddings + Additional Node Features}
- linguistic features such as sentence length, part-of-speech count, and top of three words were added as node features to Case 1.
\item \textbf{Case3 Sentence-Level Embeddings + Additional Node Features + Edge Features}
- edge features representing the cosine similarity between sentences and the order of utterances were added as edge features to Case 2
\end{itemize}

\subsection{Experimental Configurations}
Hyperparameters are crucial for model performance evaluation; therefore, we conducted experiments under the same conditions, as shown in Table \uppercase\expandafter{\romannumeral2}. Sentence embedding size represents the dimension of sentence-level embeddings; word embedding size indicates the dimension of word embeddings; input node feature denotes the dimension of node features; hidden layer signifies the dimension of hidden layers; batch size determines the number of data points used in each mini-batch during model training; learning rate controls the step size when the model updates weights using gradient descent; and dropout rate represents the proportion of nodes that are deactivated as part of a regularization technique to prevent overfitting.

\begin{table}[htbp]
    \caption{CONFIGURATION OF ECRC HYPER-PARAMETERS.}
    \begin{center}
        \begin{tabular}{lcccccccr}
            \toprule
            Parameters & Value  \\
            \midrule
            Sentence embedding size              & 1024 \\
            Word embedding size                  & 200 \\
            Input feature dimension              & 1627 \\
            Hidden layer                         & 128 \\
            Batch size                           & 32 \\
            Learning rate                        & 0.001 \\
            Dropout rate                         & 0.5 \\
            \bottomrule
        \end{tabular}
        \label{tab1}
    \end{center}
\end{table}

\section{Experimental Results}
One of the key challenges addressed in this study is the construction of diverse graph structures for conversation datasets to determine the accuracy of artificial intelligence (AI) models in predicting emotions and causality. Therefore, we first trained the ECRC model on the preprocessed ECC training dataset and evaluated its performance on the test dataset.

Table \uppercase\expandafter{\romannumeral3} presents the experimental results of the six models on the ECC dataset. A generalized performance evaluation of the ECRC model was also conducted using one English and two Korean datasets. The proposed model outperformed baseline deep learning models in multi-task learning, indicating the effectiveness of using both sentence- and word-level embeddings. Furthermore, the ECRC model with added node and edge features in the proposed model demonstrated superior performance in multi-task learning, with precision, recall, and F1-score of 74.62, 74.04, and 74.14 for emotion prediction and 75.30, 74.91, and 74.64 for causality prediction respectively, outperforming the other ECRC models. This highlights the effectiveness of the new graph structure with node and edge features. 

Table \uppercase\expandafter{\romannumeral4} displays the experimental results of the ECRC model on three multi-turn dialogue datasets. The ECC, constructed in Korean, yielded precision scores of 74.62 for emotion and 75.30 for causality. Wellness, also in Korean, resulted in precision scores of 73.44 for emotion and 73.09 for causality. The IEMOCAP yielded a precision score of 71.35 for emotion only. The experimental results demonstrate that the proposed model performs well across both Korean and English multi-turn dialogue datasets, indicating its ability to generalize performance across different language structures.

\begin{table*}[t!]
    \caption{COMPARATIVE PERFORMANCE WITH DIFFERENT MODELS, EMOTIONS, AND SITUATIONS. THE BEST-PERFORMING MODELS ARE INDICATED IN BOLD.}
    \label{sample-table}
    \vskip 0.15in
    \centering
    \resizebox{0.8\textwidth}{!}{
        \begin{tabular}{@{}lllllll@{}}
        \toprule
        \multicolumn{1}{c}{\multirow{2}{*}{Model}} & \multicolumn{3}{c}{Emotion}                                                               & \multicolumn{3}{c}{Causality}                                                             \\ \cmidrule(l){2-7} 
        \multicolumn{1}{c}{}                       & \multicolumn{1}{c}{Precision} & \multicolumn{1}{c}{Recall} & \multicolumn{1}{c}{F1-Score} & \multicolumn{1}{c}{Precision} & \multicolumn{1}{c}{Recall} & \multicolumn{1}{c}{F1-Score} \\ \midrule
        CNN                                        & 47.79     & 46.39  & 46.34    & 41.67     & 38.83  & 40.63    \\
        LSTM                                       & 52.68     & 52.14  & 51.65    & 43.74     & 39.89  & 41.71    \\ 
        Bi-LSTM                                     & 53.53     & 52.45  & 52.52    & 49.76     & 47.95  & 49.25    \\ \midrule
        ECRC(Graph)                                & 57.51     & 57.37  & 57.25    & 58.93     & 57.09  & 57.86    \\
        ECRC(Graph+Node)                           & 68.37     & 68.13  & 68.04    & 68.51     & 69.10  & 68.44    \\
        ECRC(Graph+Node+Edge)                      & \textbf{74.62}     & \textbf{74.04}  & \textbf{74.14}    & \textbf{75.30}     & \textbf{74.91}  & \textbf{74.64}                     
        \end{tabular}
    }
    \vskip -0.1in
\end{table*}

\begin{table*}[t!]
    \caption{COMPARATIVE PERFORMANCE WITH DIFFERENT DATASETS, EMOTIONS, AND SITUATIONS. FOR FAIR COMPARISON EACH MODEL WAS TRAINED AND TESTED ON THE SAME DATASET.}
    \label{sample-table}
    \vskip 0.15in
    \centering
    \resizebox{0.8\textwidth}{!}{
        \begin{tabular}{ccccccc}
        \hline
        \multicolumn{1}{l}{Datasets} & \multicolumn{2}{c}{ECC} & \multicolumn{2}{c}{Wellness} & \multicolumn{2}{c}{IEMOCAP} \\ \hline
        Model                        & Emotion    & Causality  & Emotion       & Causality    & Emotion      & Causality    \\ \hline
        \multirow{2}{*}{ECRC}        & Precision  & Precision  & Precision     & Precision    & Precision    & Precision    \\ \cline{2-7} 
                                     & \textbf{74.62}      & \textbf{75.30}      & \textbf{73.44}         & \textbf{73.09}        & \textbf{71.35}        & -           
        \end{tabular}
    }
    \vskip -0.1in
\end{table*}

\section{DISCUSSION}
The three contributions of the proposed model to context analysis are assessed. First, an effective new graph structure that incorporates both sentence-level and word embeddings was proposed. We applied sentence-level embeddings to more accurately handle polysemy and semantic changes based on context, as well as word embeddings to consider linguistic structures and grammatical features. By contrast, the baseline models primarily utilized first utterance word embeddings which have the highest weight in emotion analysis, rather than using the entire corpus. Interestingly, the baseline model's best-performing Bi-LSTM model did not outperform the ECRC model, which applies sentence-level and word embeddings. Through experiments, we demonstrated that the graph-based approach of the proposed model is effective for detecting complex patterns, analyzing the relationships between sentences, and making predictions.

Second, the proposed model demonstrated outstanding performance in multi-task learning for emotions and causality on multi-turn dialogue datasets. The deep learning approach using connectionist methods employed in previous artificial intelligence research focused on learning and generating emotional patterns through deep learning. The approach was limited to single-sentence learning and solely considered emotion classification. In contrast, our model integrates emotion and causality through multi-task learning, resulting in superior performance compared to existing ERC models, as shown in the experiments.

Third, this study is the first to analyze context using a model that applies Korean language characteristics and combines a Bi-LSTM with GCN. The proposed model adds linguistic features to nodes, including node type, embedding values, sentence length, sentence complexity, and central words. It also adds relational features to edges, including edge type and inter-sentence similarity. To extract linguistic features, we used morphological analyzers such as Mecab and NLTK; embedding models such as ELMo and word2vec; and the GCN model to learn the graph structure. Furthermore, we experimentally demonstrated that the proposed model can guarantee generalizable performance, even on multi-turn dialogue datasets in different languages, with only changes required in the morphological analyzer. The experimental results exhibited improved performance when more nodes and edge features were added.

\section{CONCLUSIONS AND FUTURE WORK}
The ECRC model demonstrated excellent performance in multi-task learning, primarily because of the node and edge features applied to the graph structure. Analyzing emotions and causality together can significantly contribute to the healthcare and medical industries, providing insight into understanding why such emotions occur.
This study only utilized text data, which limits the ability to capture the nuances and meanings in a sentence. For precise emotion analysis, implementing a model that incorporates information such as audio data frequency, amplitude, and time using multi-modal data, can potentially yield superior performance. The proposed model has a flexible structure that easily accommodates the addition of audio and image information as node and edge features, thereby enabling the development of a model more capable of considering a broader range of data types.

\vspace{11pt}

\begin{IEEEbiography}[{\includegraphics[width=1in,height=1.25in,clip,keepaspectratio]{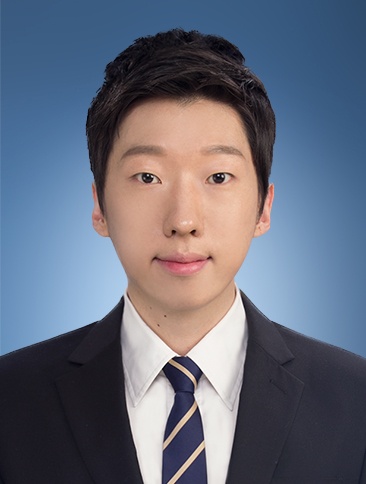}}]{Jun Koo Lee}
received the Bachelor of Science degree in Urban Engineering from Chung-Ang University, Republic of Korea, in 2016. He is currently pursuing a master's degree at Sungkyunkwan University Graduate School of Artificial Intelligence, Republic of Korea. His research interests encompass sentiment analysis, including emotion cause analysis, multi-modal sentiment analysis, and emotion recognition in conversation.
\end{IEEEbiography}

\begin{IEEEbiography}[{\includegraphics[width=1in,height=1.25in,clip,keepaspectratio]{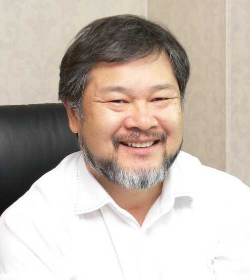}}]{Tai-Myoung Chung}
received the B.S. degree in Computer Engineering from Yonsei University in 1981, the M.S. degree in Computer Engineering from UIC in 1987, and the Ph.D. degree in Computer Engineering from Purdue University in 1995. He joined the College of Computer Science and Engineering at Sungkyunkwan University, Seoul, Korea in 1995, where he is currently an Emeritus Professor. He contributed to the ICT Advisory Committee for Rwanda in 2010. In 2011, he became the President of the Korea Information Processing Society, where he made significant advancements in the field. In 2007, he became the President of the CPO Forum. In 2020, he founded Hippo T\&C and became its CEO, leading the company to excel in digital healthcare. His research interests are centered around SDN-based security, IoT management, and mobile computing security.
\end{IEEEbiography}

\begin{IEEEbiography}[{\includegraphics[width=1in,height=1.25in,clip,keepaspectratio]{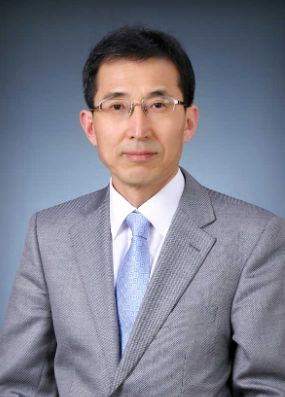}}]{Moon-Hyun Kim}
received the B.S. degree in Electronic Engineering from Seoul National University in 1978, the M.S. degree in Electrical Engineering from KAIST, Korea, in 1980, and the Ph.D. degree in Computer Engineering from the University of Southern California in 1988. He joined the College of Information and Communication Engineering, Sungkyunkwan University, Seoul, Korea in 1988, where he is currently a Emeritus Professor. From 2021,he joined HippoT\&C as a research director. In 1995, he was a Visiting Scientist at the IBM Almaden Research Center, San Jose, California. In 1997, he was a Visiting Professor at the Digital Neural Network Laboratory of Princeton University, Princeton, New Jersey. His research interests include machine learning , artificial intelligence, and pattern recognition.
\end{IEEEbiography}

\end{document}